\pdfoutput=1

\documentclass[11pt]{article}
\makeatletter
\def\input@path{{./latex/}{./}}
\makeatother

\usepackage[final]{acl}

\usepackage{times}
\usepackage{latexsym}

\usepackage[T1]{fontenc}

\usepackage[utf8]{inputenc}

\usepackage{microtype}

\usepackage{inconsolata}

\usepackage{graphicx}
\usepackage{booktabs} 
\usepackage{multirow}
\usepackage{algorithm}
\usepackage{algpseudocode}

\usepackage{enumitem}
\usepackage{graphicx}
\usepackage{amsmath}
\usepackage{amssymb}
\usepackage{mathtools}
\usepackage{amsthm}
\usepackage{bm}
\usepackage{float}
\usepackage[export]{adjustbox}
\usepackage{soul}
\usepackage{booktabs}

\usepackage{caption}
\usepackage{multirow}
\usepackage{makecell}
\usepackage{wrapfig}
\usepackage[utf8]{inputenc}
\usepackage[T1]{fontenc}

\usepackage{tcolorbox}

\newcommand{\nn}{\texttt{<\textbackslash n\textbackslash n>}}

\title{Word Salad Chopper: Reasoning Models Waste\\ A Ton Of Decoding Budget On Useless Repetitions, Self-Knowingly}

\author{
\textbf{Wenya Xie\textsuperscript{1}\footnotemark[1]}, 
\textbf{Shaochen (Henry) Zhong\textsuperscript{2}\footnotemark[1]}, 
\textbf{Hoang Anh Duy Le\textsuperscript{2}},\\
\textbf{Zhaozhuo Xu\textsuperscript{3}}, 
\textbf{Jianwen Xie\textsuperscript{4}},
\textbf{Zirui Liu\textsuperscript{1}},\\
\textsuperscript{1}University of Minnesota \quad
\textsuperscript{2}Rice University \quad\\
\textsuperscript{3}Stevens Institute of Technology \quad
\textsuperscript{4}Lambda, Inc \quad
}

\begin{document}

\maketitle
\let\svthefootnote\thefootnote
\newcommand\freefootnote[1]{%
  \let\thefootnote\relax%
  \footnotetext{#1}%
  \let\thefootnote\svthefootnote%
}
\freefootnote{* Equal contribution.}

\begin{abstract}


Large Reasoning Models (LRMs) are often bottlenecked by the high cost of output tokens. We show that a significant portion of these tokens are useless self-repetitions — what we call \textit{“word salad”} — that exhaust the decoding budget without adding value. Interestingly, we observe that LRMs are self-aware when trapped in these loops: the hidden states of \nn{} tokens trailing each reasoning chunk exhibit patterns that allow us to detect word salad behavior on-the-fly via a single-layer linear classifier. Once detected, a simple chop appended by a straightforward regeneration prompt yields substantial length savings with minimal quality loss. Our work offers WordSaladChopper (WSC) — a lightweight, turnkey component for LRM that is minimally invasive to its reasoning trajectory by only removing semantically redundant tokens. Given its low overhead, strong savings, and the lack of semantic value of word salad tokens, \textbf{we believe it is not too far-fetched to argue that WSC — or a similar component — is a must-have for all LRM applications with user experience in mind. }Our code is publicly available at \url{https://github.com/wenyaxie023/WordSaladChopper}.

\end{abstract}

\section{Introduction}
\vspace{-0.5em}
\label{sec_intro}

\begin{figure*}
    \centering
    \includegraphics[width=\textwidth]{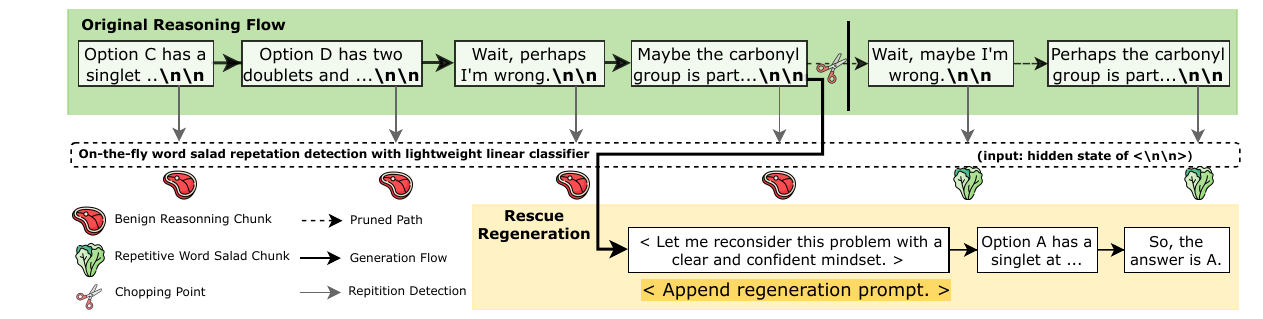}
    \caption{General workflow of WordSaladChopper. 1) \textbf{Detect}: We allow the reasoning model to freely generate, following its original reasoning flow. Meanwhile, we classify the hidden state of each chunk's trailing \nn{} token using our trained linear classifier in an on-the-fly manner; 2) \textbf{Chop}: Once a chopping point is reached — in this case, it is defined by having two consecutive \textit{word salad chunks} detected — we truncate the generation to the left of it; 3) \textbf{Regenerate}: We append a regeneration prompt with constant budget, allowing the model to complete its answer by its own via \texttt{<eos>} or until the new budget is fully expensed.}
    \vspace{-5mm}
    \label{fig:method}
    
\end{figure*}

Despite the drastic boost in performance over their non-reasoning counterparts, one innate issue of LRMs is that they essentially trade more decoded tokens for capabilities. However, \textbf{a prolonged decoding section is among the most expensive operations a Large Language Model (LLM) can experience} due to compute, memory, and scheduling challenges. For instance, OpenAI o3 charges \$10/\$40 per one million of input/output tokens,\footnote{\url{https://openai.com/api/pricing/}} a striking 4$\times$ difference between decoding and prefill. Despite the high cost of long thinking traces, a less well-known and rarely quantified fact~\cite{li2025small, yeo2025demystifying} is that \textbf{LRMs tend to waste an enormous amount of decoding budget, simply by repeating themselves} verbatim, with slight variations, or engaging in endless enumeration of cases until all budget has been expensed (see examples at Appendix~\ref{sec:case-study}) — we refer to such behavior as \textbf{\textit{Word Salad}}, a term often used to mock public spokespersons for giving long-winded, jargon-filled responses that ultimately lack substance or clear meaning. The ``Original'' column in Table~\ref{tab_salad_token_count} shows that when answering GPQA-Diamond \citep{rein2024gpqa}, we observe 55\%+ of tokens generated by \texttt{DeepSeek-R1-Distill} models are marked as ``word salad tokens,'' where they do not add value from a semantic standpoint.\footnote{For better flow, we refer readers to Section~\ref{sec_ob_heavy_contributor} for the technical definition of \textit{word salad tokens}. Intuitively, one can understand them as a rough catch-all for different kinds of verbatim or non-verbatim repetitive behaviors that do not provide much value from a semantic standpoint.}

\begin{table}[H]
\centering
\vspace{-3mm}

\caption{Percentage of \textit{word salad tokens} in answering GPQA-Diamond. 55\%+ of the budget has been wasted.}
\label{tab_salad_token_count}

\vspace{-2mm}

\resizebox{1\linewidth}{!}{
\begin{tabular}{l|cc}
\toprule

\textbf{Model}  & \textbf{Original} & \textbf{After Chop} \\

\midrule

\texttt{DeepSeek-R1-Distill-Qwen-1.5B} & 63.37 & 5.29 \\
\texttt{DeepSeek-R1-Distill-Qwen-7B} & 61.92 & 4.23 \\
\texttt{DeepSeek-R1-Distill-Llama-8B} & 56.60 & 5.60 \\

\bottomrule
\end{tabular}
 }

\vspace{-4.5mm}
\end{table}

Naturally, making such thinking sections shorter while preserving answer quality has become a major goal of the efficiency community. In fact, many works have emerged in a short period, forming a new subfield of \textit{long-to-short (L2S)}; with some of the most effective L2S methods often requiring training intervention \citep{sui2025stop_overthink_survey, wang2025reason_eco_survey, liu2025efficient_lrm_survey}. While effective, with major parameter updates, such training-based L2S methods surely introduce a rather aggressive ``invasion'' into the original reasoning trajectory of LRMs, where the side effects remain largely unknown. Moreover, such methods typically do not stack well with one another, as different training recipes often demand intrinsically conflicting operations.\quad Instead, in this work, we explore whether it is possible to advance efficient reasoning in a \ul{turnkey} and \ul{minimally invasive} manner, just by reducing the \textit{word salad} behavior — as such salad tokens are likely universally agreed to be redundant, if not at all useless, from a semantic standpoint.

Surprisingly, we find that the \textbf{model is actually self-aware when it is trapped in such ``word salad'' loops} — specifically, the hidden states of \nn{} tokens at the end of each reasoning chunk show distinguishable patterns when the model is trapped versus when it is not. Leveraging this observation, we train a lightweight linear classifier that runs on-the-fly to detect this word salad behavior. Once detected, a simple chop and regeneration prompt yields significant length savings with minimal quality loss — e.g., \textbf{the chopping would immediately reduce up to 92\% of word salad tokens} in \texttt{DeepSeek-R1-Distill-Qwen-7B} when undergoing GPQA-Diamond (Table~\ref{tab_salad_token_count}). In summary, our main contributions are as follows:

\begin{itemize}[leftmargin=*, noitemsep, topsep=0pt]
    \item \textbf{Comprehensive investigation of LRM word salad behavior.} To the best of our knowledge, we are the first to systematically study the general repetition phenomenon in LRM reasoning traces, identifying its key characteristics, persistence, and its robustness to existing reputation penalties.
    
    \item \textbf{Empirical evidence that LRMs are self-aware when trapped in word salad loops.} We show that the hidden states of \nn{} tokens carry distinct signals when the model is stuck in word salad loops versus when it is reasoning normally — revealing a hidden opportunity for detection and intervention.
    
    \item \textbf{A lightweight, turnkey, minimally invasive component for all LRM applications.} We propose a specially-trained linear classifier that runs on-the-fly without retraining or architectural modification on the LRM end. Once word salad behavior is detected, a chop-then-regenerate routine significantly reduces output length with minimal reasoning quality degradation.
\end{itemize}

\noindent In this work, we aim to deliver our following messages clearly and quickly: 1) Word salad is an overlooked but severe issue present across likely all LRMs. It offers no benefit yet consumes an atrocious amount of decoding budget; and 2) LRMs are self-aware of such behavior, where on-the-fly detection and intervention is possible. We believe any LRM-serving application should consider adopting our component — or something similar — as an almost-free-lunch solution for immediate cost savings and latency improvements. Due to page limitations and lack of tightly relevant art, we refer the reader to Appendix~\ref{app_related_works} for Related Work discussions.


\section{Observations}
\vspace{-0.5em}
\label{sec_moti}
In this section, we outline four empirically supported observations of LRM word salad behavior.

\subsection{A Heavy Contributor of Long Thinking is Word Salad-like Self-Repetitions}
\vspace{-0.5em}
\label{sec_ob_heavy_contributor}

Much of the contribution of our work depends on whether there truly exists a significant amount of word-salad-like self-repetitions within LRM's reasoning traces. Defining such behavior demands carefulness, as LRMs typically do not exhibit strictly verbatim repetitions, rendering rule-based methods not applicable. To achieve an accurate yet simple flagging, we employ an embedding model $E$. Then, for a given trace $T$, we first chunk $T$ into different chunks based on some common delimiter — in this case, \nn{} — so we'd have $T = c_1 \oplus c_2 \oplus \dots \oplus c_n$ where $c_i$ represents the $i$-th chunk of $T$ and $\oplus$ represents concatenation. A chunk $c_i$ is considered a ``word salad chunk'' if $E(c_i, c_j) \geq \theta$ for $j = \{1, 2, \dots, i-1\}$, where $\theta$ is a similarity threshold.\footnote{In practice, we set $\theta=0.99$ \& $E=$ \texttt{all-MiniLM-L6-v2}.} Namely, $c_i$ is flagged as a word salad chunk if it is highly similar to a previous chunk $c_j$ within the thinking trace $T$, per the embedding model $E$. We consider all tokens within a word salad chunk as word salad tokens. 

\begin{table}[ht]
\vspace{-1mm}
\centering
\caption{Percentage of \textit{word salad chunks} within reasoning traces. Result are presented as  (temp $\tau$ = 0.0, 0.6).}
\vspace{-2mm}
\label{tab_repetition_rate}
\resizebox{\linewidth}{!}{
\begin{tabular}{lc|c|c|c}

\toprule
 
\textbf{Model}   &    \textbf{GSM8K} &    \textbf{MATH-500} &    \textbf{AIME25} &    \textbf{GPQA-Diamond}  \\

\midrule

Qwen-1.5B & (51.2, 37.4)   &    (62.9, 10.6)     &    (77.5, 18.7) &    (87.7, 42.4) \\
Qwen-7B   & (23.9, 8.1)    &    (45.4, 10.9)     &    (52.1, 10.9) &    (72.7, 25.3) \\
Llama-8B  & (35.0, 8.3)    &    (53.1, 10.5)     &    (62.9, 13.6) &    (60.1, 18.0) \\

\bottomrule
\end{tabular}
}
\vspace{-4mm}
\end{table}

Table~\ref{tab_repetition_rate} indicates that such word salad chunks indeed occupy a non-trivial presence in the reasoning traces. We additionally note that, unless otherwise specified, all reported models are of \texttt{DeepSeek-R1-Distill} series with temp $\tau=0$.

\subsection{Once Word Salad Happens, LRMs are Unlikely to Get Out on Their Own}
\vspace{-0.4em}
\label{sec_ob_salad_stop}

One unique characteristic of word salad that would result in a poor user experience is that once the model triggers word salad, it is unlikely to untrap itself. Thus, the model will most likely be trapped in such word salad loops until all the decoding budget has been fully expensed. We refer to this boundary as the \textit{chopping point} (Table~\ref{tab:repetition_ratio_boundary}).

\begin{table}[ht]
\vspace{-2mm}
\centering
\caption{Percentage of \textit{word salad chunks} before / after the \textit{chopping point}.}
\vspace{-3mm}
\label{tab:repetition_ratio_boundary}
\resizebox{\linewidth}{!}{
\begin{tabular}{lc|c|c|c}

\toprule
 
Model   &    \textbf{GSM8K} &    \textbf{MATH-500} &    \textbf{AIME25} &    \textbf{GPQA-Diamond}  \\

\midrule

\multicolumn{5}{c}{\textbf{$\tau$ = 0.0}} \\

\midrule

\texttt{Qwen-1.5B} & 2.08 / 98.00   &    9.48 / 94.91     &    11.68 / 99.05 &    17.19 / 96.93 \\
\texttt{Qwen-7B} & 1.21 / 98.30   &    6.59 / 89.63     &    10.03 / 81.82 &    13.13 / 95.63 \\

\midrule

\multicolumn{5}{c}{\textbf{$\tau$ = 0.6}} \\

\midrule

\texttt{Qwen-1.5B} & 2.75 / 97.21   &    8.23 / 51.35     &    8.95 / 60.07 &    8.84 / 93.92 \\
\texttt{Qwen-7B} & 0.34 / 77.32   &    2.30 / 21.80     &    3.10 / 13.79 &    1.93 / 42.81 \\

\bottomrule
\end{tabular}
}
\vspace{-3mm}
\end{table}

Needless to say, this presents a catastrophic issue to users, as an ideally much shorter thinking section is now maximized with useless repetitions. So the user is essentially paying the maximum cost for a (likely) wrong answer, while enduring the longest end-to-end latency. In practice, we find that \texttt{Qwen-1.5B} often requires a much longer runtime than its \texttt{7B} counterpart, for the exact reason that it is maximizing its decoding budget a lot more often with word salad chunks. This goes against the main drive of using smaller LRM in the first place.

\subsection{Such Kind of ``Word Salad'' Behavior is Hard to Address with Existing Means.}
\vspace{-0.5em}
\label{sec_ob_hard_existing}

The previous two observations demonstrated the prevalence and severity of word salad. However, this is really only an issue if it cannot be trivially addressed via existing detection methods or various available repetition penalty designs. Given that our word salad detection, as described in Section~\ref{sec_ob_heavy_contributor}, relies on leveraging an embedding model $E$ to compute pairwise chunk similarities, the pipeline itself naturally serves as a mechanism for identifying word salad behavior. However, this approach is far from efficient enough to be deployed on-the-fly, as it incurs a complexity of $\Theta\bigl(n^{2})$ for $n$ chunks. Even with cashing, each operation requires fully passing one chunk through $E$, which is infeasible to be deployed on-the-fly. 

One alternative avenue is to employ existing \href{https://platform.openai.com/docs/api-reference/completions}{decoding penalties}, such as repeat~\cite{keskar2019ctrl}, presence, and frequency penalties. Unfortunately, those penalties introduce much randomness to the correctness of LRMs, often negatively. Results from Table~\ref{tab_penalty_diff_settings} suggest they are too aggressive in their invasions of the reasoning trajectory of LRMs, and therefore too volatile to be usable.

\begin{table}[ht]
\centering
\vspace{-2mm}
\caption{Task performance w/ penalties ($\tau$ = 0.6)}
\vspace{-2mm}
\label{tab_penalty_diff_settings}
\resizebox{\linewidth}{!}{
\begin{tabular}{lc|c|c|c}

\toprule
 
\textbf{Decoding Setting}   &    \textbf{GSM8K} &    \textbf{MATH-500} &    \textbf{AIME25} &    \textbf{GPQA-Diamond}  \\

\midrule

Vanilla & 89.76   &    90.80     &    37.92 &    43.43 \\
\midrule
Repeat Penalty & 86.05   &    87.20     &    25.83 &    49.49 \\
Presence Penalty & 89.61  &    89.80     &    41.67 &    48.48 \\
Frequency Penalty & 78.54   &    43.80     &    13.33 &    36.87 \\

\bottomrule
\end{tabular}
}
\vspace{-3mm}
\end{table}

\vspace{-0.5em}
\subsection{Models are Self-Aware when it is Trapped in Word Salad Loops}
\label{sec_ob_self_aware}
\vspace{-0.5em}

We, rather surprisingly, find that LRMs are self-aware when they are trapped in word salad loops. Specifically, we find that it is possible for us to train a simple linear classifier — with special data curation and training recipe detailed in Section~\ref{sec_pm_lc_train} — to distinguish the hidden state of trailing \nn{} token of word salad chunks versus benign reasoning chunks. The lightweightness of this linear classifier opens the door for on-the-fly detection, where we can effectively intervene with different operations to address models trapped in word salad loops. Table~\ref{tab:probe_performance} supports the effectiveness of this classifier.

\begin{table}[ht]
\centering
\vspace{-2mm}
\caption{Classifier performance on \textit{word salad chunks} detection with \texttt{Qwen-7B}. Results as (Acc. / AUROC).}
\vspace{-2mm}
\label{tab:probe_performance}
\resizebox{\linewidth}{!}{
\begin{tabular}{lc|c|c|c}

\toprule
 
Temp   &    \textbf{GSM8K} &    \textbf{MATH-500} &    \textbf{AIME25} &    \textbf{GPQA-Diamond}  \\

\midrule

$\tau$ = 0.0 & 92.72 / 98.63   &    92.31 / 95.95     &    89.77 / 95.84 &    93.52 / 97.89 \\
$\tau$ = 0.6 & 91.42 / 96.22   &    88.14 / 95.26     &    77.96 / 80.15 &    93.80 / 96.96 \\

\bottomrule
\end{tabular}
}
\vspace{-4mm}
\end{table}

\vspace{-0.5em}
\section{Proposed Method}
\vspace{-0.3em}
\newcommand{\minus}{\raisebox{.2ex}{\scalebox{0.8}{$-$}}}
\label{sec_proposed_method}



\subsection{Training a Lightweight Linear Classifier as the Word Salad Chunk Detector}
\vspace{-0.3em}
\label{sec_pm_lc_train}

Based on observations from Section~\ref{sec_ob_heavy_contributor} and \ref{sec_ob_salad_stop}, we are aware that chunks after the \textit{chopping point} are primarily word salad chunks. Thus, it is practically sensible to mark all chunks after these chopping points as word salad chunks — even if some of them are not by definition of Section~\ref{sec_ob_heavy_contributor} — as stopping generation at the chopping point is reasonable.

\paragraph{Data Curation} Following this design principle, we collect \textbf{1,000 seed thinking traces by feeding the s1~\citep{muennighoff2025s1} questions to each model tested}. Adopting the similar methodology from Section~\ref{sec_ob_heavy_contributor}, we first chunk each thinking trace $T$ as $n$ chunks by $T = \{c_1, c_2,\dots, c_n\}$ by \nn{}.\footnote{To clarify, $n$ is not a constant set by us, but naturally derived from the number of \nn{} in $T$.} Then, we label chunk $c_i$ as ``word salad chunk'' (say label \texttt{1}) if $E(c_i, c_j) \geq \theta$ for $j < i$, where $\theta$ is a similarity threshold set to 0.99; otherwise, $c_i$ is labeled as a ``benign reasoning chunk'' (say with label \texttt{0}). Additionally, to avoid undesired long range dependency (labeling a chunk as word salad because a much, much earlier chunk is considered similar to it), we limited $(j-i) \leq 100$. We then identify the chunk of the earliest chopping point $c_t$ within this labeled $T$, where $k-1$ consecutive chunks of $c_t$ are all labeled as word salad chunks. We then relabel all chunks before $c_t$ as label \texttt{0} and all chunks including and after $c_t$ as \texttt{1}.

\paragraph{Training Recipe} With this relabeled data collected, we collect the output of the final transformer block of each \nn{} from models, along with their binary labels, to train a linear classifier consisting of a fully-connected layer, as detailed in Appendix~\ref{app:training details}. We emphasize that we essentially only ``pretrain'' this lightweight linear classifier once per each model on our s1-curated data, where all reasoning evaluation results are collected on unseen data with no finetuning involved.

\vspace{-0.5em}
\subsection{Detect, Chop, then Regenerate} 
\label{pm_detect}
\vspace{-0.3em}

Due to space limits, we refer readers to Figure~\ref{fig:method} for the WordSaladChopper workflow. As supporting evidence, Table~\ref{tab:probe_performance} shows that the linear classifier is extremely accurate in detecting the word salad chunks; yet Table~\ref{tab:trimmed_acc} demonstrates that the regeneration prompt helps recover the task accuracy lost from brute-force chopping.

\begin{table}[ht]
\vspace{-3mm}
\centering
\caption{\underline{Original}/Chopped/\textit{Regenerated} Acc. for \texttt{Qwen-7B} at $\tau=0.6$}
\vspace{-3mm}
\label{tab:trimmed_acc}
\resizebox{\linewidth}{!}{
\begin{tabular}{c|c|c|c}
\toprule
 \textbf{GSM8K} & \textbf{MATH-500} & \textbf{AIME25} & \textbf{GPQA-Diamond} \\
\midrule
 \underline{89.76} / 78.24 / \textit{89.69} & \underline{90.8} / 83.2 / \textit{89.60} & \underline{37.92} / 29.17 / \textit{37.92} & \underline{43.43} / 42.93 / \textit{43.43} \\
\bottomrule
\end{tabular}
}
\vspace{-4mm}
\end{table}

\vspace{-0.5em}
\section{Experiments and Discussion}
\vspace{-0.5em}
\label{sec_sexp}

\begin{table}[H]
\vspace{-2.5mm}
\centering
\caption{End-to-end task performance of WSC w/ $\tau=0$ in terms of task accuracy and length compression. \small{(AIME25 is omitted here as the variance can be extreme w/ $\tau=0$, where only one pass of 30 questions is possible.)}}
\vspace{-2mm}
\label{tab:e2e_t0}
\resizebox{\linewidth}{!}{
\begin{tabular}{lcc|cc|cc}

\toprule
\multirow{1}{*}{\textbf{Setting}}  &   \multicolumn{2}{c}{\textbf{GSM8K}}   &   \multicolumn{2}{c}{\textbf{MATH-500}}   &   \multicolumn{2}{c}{\textbf{GPQA-Diamond}}\\

\cmidrule{2-7}
 
& Acc.    &    Len.    &    Acc.    &    Len.    &    Acc.    &    Len.    \\

\midrule

\multicolumn{7}{c}{\textbf{Qwen-1.5B}} \\
\midrule
Original & 82.03 & 1904 & 72.20 & 8126 & 32.83 & 23449 \\ 
\multirow{2}{*}{WSC (Ours)} 
& 82.64 & 1082 & 72.60 & 4253 & 31.82 & 10004 \\
&\textcolor{gray}{\small ↑0.61} 
&\textcolor{gray}{\small ↓43.19\%} 
&\textcolor{gray}{\small ↑0.40} 
&\textcolor{gray}{\small ↓47.66\%} 
&\textcolor{gray}{\small ↓1.01} 
&\textcolor{gray}{\small ↓57.34\%} \\
\midrule

\multicolumn{7}{c}{\textbf{Qwen-7B}} \\
\midrule
Original & 89.99 & 758 & 87.60 & 4925 & 44.95 & 12974 \\
\multirow{2}{*}{WSC (Ours)} 
& 90.45 & 567 & 86.80 & 3399 & 42.42 & 6027 \\
&\textcolor{gray}{\small ↑0.46} 
&\textcolor{gray}{\small ↓25.23\%} 
&\textcolor{gray}{\small ↑0.20} 
&\textcolor{gray}{\small ↓31.00\%} 
&\textcolor{gray}{\small ↓2.53} 
&\textcolor{gray}{\small ↓53.55\%} \\
\midrule

\multicolumn{7}{c}{\textbf{Llama-8B}} \\
\midrule
Original & 85.60 & 894 & 79.20 & 5556 & 38.89 & 11969 \\
\multirow{2}{*}{WSC (Ours)} 
& 85.67 & 667 & 80.4 & 3684 & 38.89 & 7292 \\
&\textcolor{gray}{\small ↑0.07} 
&\textcolor{gray}{\small ↓25.40\%} 
&\textcolor{gray}{\small ↑1.20} 
&\textcolor{gray}{\small ↓33.69\%} 
&\textcolor{gray}{\small ↑0.00} 
&\textcolor{gray}{\small ↓39.07\%} \\
\bottomrule
\end{tabular}
}
\vspace{-3mm}

\end{table}

\begin{table}[H]
\centering
\vspace{-4mm}
\caption{End-to-end task performance of WSC w/ $\tau = 0.6$. (\small AIME25 results are averaged over 8 passes.)}
\vspace{-2mm}
\label{tab:e2e_t0.6}

\resizebox{\linewidth}{!}{

\begin{tabular}{lcc|cc|cc|cc}
\toprule
\multirow{1}{*}{\textbf{Setting}}  &   \multicolumn{2}{c}{\textbf{GSM8K}}   &   \multicolumn{2}{c}{\textbf{MATH-500}}   &   \multicolumn{2}{c}{\textbf{AIME25}} &   \multicolumn{2}{c}{\textbf{GPQA-Diamond}}\\

\cmidrule{2-9}
 
&    Acc.    &    Len.    &    Acc.    &    Len.    &    Acc.    &    Len.    &    Acc.    &    Len. \\

\midrule

\multicolumn{9}{c}{\textbf{Qwen-1.5B}} \\
\midrule
Original & 82.56 & 1012 & 81.60 & 4485 & 21.67 & 16462 & 35.86 & 7790 \\
\multirow{2}{*}{WSC (Ours)} 
& 83.02 & 818 & 80.40 & 4065& 21.67& 13591 & 35.35 & 5708\\ 
&\textcolor{gray}{\small↑0.46} &
\textcolor{gray}{\small ↓19.20\%} 
&\textcolor{gray}{\small ↓1.23} 
 &\textcolor{gray}{\small ↓9.38\%} 
&\textcolor{gray}{\small ↑0.00} 
 &\textcolor{gray}{\small ↓17.44\%} 
&\textcolor{gray}{\small ↓0.45} 
 &\textcolor{gray}{\small ↓26.73\%} \\
\midrule

\multicolumn{9}{c}{\textbf{Qwen-7B}} \\
\midrule
Original & 89.76 & 565 & 90.80 & 3597 & 37.92 & 15305 & 43.43 & 6201 \\
\multirow{2}{*}{WSC (Ours)} 
& 89.99 & 545 & 90.40 & 3215 & 36.25 & 12239 & 43.43 & 5345 \\
&\textcolor{gray}{\small ↑0.23} 
&\textcolor{gray}{\small ↓3.44\%} 
&\textcolor{gray}{\small ↓0.40} 
&\textcolor{gray}{\small ↓10.62\%} 
&\textcolor{gray}{\small ↓1.67} 
&\textcolor{gray}{\small ↓20.03\%} 
&\textcolor{gray}{\small ↑0.00} 
&\textcolor{gray}{\small ↓13.81\%} \\
\midrule

\multicolumn{9}{c}{\textbf{Llama-8B}} \\
\midrule
Original & 85.75 & 650 & 83.60 & 3899 & 28.75 & 14358 & 44.44 & 7061 \\
\multirow{2}{*}{WSC (Ours)} 
& 85.67 & 650 & 83.8 & 3641 & 29.16 & 13768 & 44.44 & 6604 \\
&\textcolor{gray}{\small↓0.08} 
&\textcolor{gray}{\small ↓1.32\%} 
&\textcolor{gray}{\small ↑0.20} 
&\textcolor{gray}{\small ↓6.60\%} 
&\textcolor{gray}{\small ↑0.42} 
&\textcolor{gray}{\small ↓4.11\%} 
&\textcolor{gray}{\small ↑0.00} 
&\textcolor{gray}{\small ↓6.46\%} \\
\bottomrule
\end{tabular}
}
\vspace{-3mm}
\end{table}







\paragraph{Result Discussion} Table~\ref{tab:e2e_t0} and \ref{tab:e2e_t0.6} showcased the effectiveness of our method, where we shall observe WordSaladChoper is capable of yielding similar reasoning benchmark performance to the original model but with reduced length. We emphasize that this is achieved with negligible overhead, as once the linear classifier is trained, the inference of this linear classifier consists of passing the hidden state of just one \nn{} token for each chunk. Given the fact that this linear classifier is so lightweight, its wall-clock runtime is exponentially quicker than decoding a full chunk in an LRM, making the overhead nicely hidden from an LRM inference perspective (see Appendix~\ref{app:latency} for details).

\vspace{-0.5em}
\section{Conclusion}
\vspace{-0.5em}
Our work investigates the phenomenon of word salad behavior in LRM and introduces a lightweight, turnkey, minimally invasive way to reduce such useless budget wasting.

\newpage
\section*{Limitations}

While our WordSaladChopper successfully curbs the onset of repetition and maintains answer completeness through fixed-budget regeneration, we observe that certain generations still lapse into repetitive loops even after the rescue regeneration phase. This suggests that future work will require more robust and adaptive interventions to effectively disengage the model from such failure modes.

\textbf{We emphasize that our work is not to present an end-to-end solution that addresses the general long-to-short task of efficient reasoning; rather, we intend to highlight the severity of word salad behaviors and present a new avenue for effective LRM control and usage.} Our regeneration prompt is presented as the most straightforward way to accompany word salad reduction, and there sure can be more sophisticated ways to deal with such post-chopping operations. For instance, one can explore the following strategies.

\begin{itemize}[leftmargin=*]
\item Grant the model a small regeneration budget after the regeneration prompt (our approach in this work). So even if it repeats, it will max out soon.
\item Continuously apply WordSaladChopper for more chopping and more regenerations.
\item Force append an end-of-think token and compel the model to output an answer on the spot. This can be combined with strategies above — giving the model a limited regeneration budget, letting it keep thinking, chopping and regenerating if necessary; then, when the budget is nearly or fully expended, forcing it to conclude and provide a short answer.
\end{itemize}

\noindent We made the decision (of not exploring sophisticated end-to-end solutions) consciously because we truly believe a WSC-like component can be a must-have turnkey addition to any LRM serving system — as no one wants to waste decoding budget on useless repetitions. So, how it is integrated into different systems will naturally demand variations. 

Further, \textbf{it is our honest belief that many efficient reasoning methods appear effective partly because current reasoning evaluation benchmarks have much room for improvement.} Should we develop more comprehensive evaluation suites \citep{gema2025inverse, huan2025does} — which we surely will in the future  — we expect to see many efficient reasoning methods fail, or behave much differently than their vanilla LRM counterparts.\footnote{And there is nothing wrong with that, just the typical trade-offs and good progression of science.} \textbf{For this reason, we want to make our approach as faithful to the original reasoning trajectory of the LRM as possible, as this is failproof to benchmark deficiency.} We therefore keep the operations after the chop simple and straightforward — as there is no useful ``reasoning trajectory ground truth'' to adhere to once the model is already trapped in a word salad loop.

Last, we want to highlight that since our Chopper requires model-specific training, it is possible that its performance may vary under different model-task combinations. We kindly ask our end users to practice caution when adopting our method.

\section*{Ethical Considerations}

We do not believe our work is applicable to ethical review, though our work does interfere with the original output of the model, where end users should treat its output with care.

\section*{Acknowledgments}
We gratefully acknowledge the support of Lambda, Inc. for providing the compute for this project. The work of Zhaozhuo Xu and Zirui Liu is supported by NSF 2450524. Zhaozhuo Xu is also supported by NSF 2451398. Wenya Xie is supported in part by the Data Science Initiative (DSI) Fellowship at the University of Minnesota.



\newpage
\appendix
\section{Related Works}
\label{app_related_works}

\paragraph{Training-based Long to Short (L2S)}
Large reasoning models often produce lengthy chain-of-thoughts due to the refinement of intermediate reasoning, inflating latency and cost. A series of post-training approaches~\citep {yan2025inftythink, munkhbat2025self} teach models to reach correct answers with fewer tokens by constructing more concise training data. TokenSkip~\cite{xia2025tokenskip}, SPIRIT-FT~\cite{cui-etal-2025-stepwise}, Coconut~\cite{hao2024training}, and CCoT~\cite{nayab2024concise}, which shorten reasoning traces via finetuning or latent-space supervision. These methods are effective but require re/post-training and are sometimes tied to specific architectures. Reinforcement learning approaches~\citep{aggarwal2025l1, hou2025thinkprune} explicitly give short length as a reward to reduce response length. While effective, all of these approaches require additional finetuning (either on LRMs or from a non-thinking model) and cannot directly function upon off-the-shelf LRMs. 

Our main reservation about such kinds of approaches is, finetuning heavily perturbs the original reasoning trajectory of the LRMs. Although most L2S literature claims that they experience minimal performance degradation, it is our honest belief that many efficient reasoning methods appear effective partly because current reasoning evaluation benchmarks have much room for improvement. Should we develop more comprehensive evaluation suites — which we surely will in the future — we expect to see many efficient reasoning methods fail, or behave much differently than their vanilla LRM counterparts (and there is nothing wrong with that, just the typical trade-offs and good progression of science). For this very reason, we want to make our approach as faithful to the original reasoning trajectory of the LRM as possible, as this is failproof to benchmark deficiency. We therefore keep the operations after the chop simple and straightforward — as there is no useful ``reasoning trajectory ground truth'' to adhere to once the model is already trapped in a word salad loop. 


\paragraph{On the fly and training-free intervention}
Rather than additional finetuning, many methods attempt lightweight control during inference. Prompt-based strategies like TALE~\cite{han-etal-2025-token} and Sketch-of-Thought~\cite{aytes2025sketch} control generation budgets via prompt engineering, but they rely on accurate length estimation and often struggle with complex reasoning. Difficulty-aware budgeting approaches such as DSC~\cite{wang-etal-2025-make}  and Dynasor~\cite{nayab2024concise} dynamically allocate compute based on estimated query difficulty or model confidence. While they share similarities with WSC in adapting decoding, they operate at the query level, whereas WSC monitors intra-sequence reasoning dynamics.

A second line of work directly manipulates the decoding process. ESC~\citep{li2024escape} dynamically stops the sampling process when a local observation window reaches a low-entropy state, while DEER~\citep{yang2025dynamic} exploits hidden-state transitions to plant new reasoning paths upon high provisional confidence. \citet{zhang2025reasoning} trains a linear probe on hidden states to predict correctness and halt decoding early. Additionally, some methods apply decoding-time penalties to discourage repetitive outputs, such as repeat penalty~\citep{keskar2019ctrl} and frequency and presence penalties\footnote{See \url{https://platform.openai.com/docs/api-reference/completions} for details}. 
However, these methods can alter the model's original reasoning trajectory and may damage overall performance.
In contrast, our method focuses on identifying the onset of repetitive behavior — an orthogonal dimension of redundancy — and intervenes only to prevent pretentious loops, thereby preserving the model’s full reasoning capabilities.

\paragraph{LRM repetition}

We emphasize that repetition (and, by extension, overthinking) in LLMs/LRMs has received increasing attention, where our work is certainly not the first to notice such repetition behaviors — evident from the long-standing repetition penalties highlighted and featured in our Section~\ref{sec_ob_hard_existing}. Here, we feature several more modern studies regarding LRM repetition.

\citet{wang2025thoughts} provides a valuable analysis of overthinking behaviors and proposes a self-training-based finetuning approach to simplify reasoning trajectories. Its link to repetition appears mainly in Section 2.3, where the authors observe that later solutions sometimes repeat earlier ones and therefore promote solution diversity. Ultimately, \citet{wang2025thoughts} is a typical L2S method that utilizes a compound finetuning approach to encourage several desirable reasoning behaviors (not just repetition reduction) by finetuning on the model’s self-generated data. WSC differs from it by providing inference-time repetition detection with negligible overhead. \textbf{To the best of our knowledge, no prior work offers on-the-fly detection of repetition in LRMs, and this lightweight capability makes WSC a turnkey drop-in for most reasoning pipelines}, including \citet{wang2025thoughts}.

\citet{yao-etal-2025-understanding} leverages pretrained Sparse Autoencoders (SAEs) to pinpoint layer-specific ``repetition features,'' then performs activation patching to damp those features and lower the repeat score. The method is not lightweight enough for true on-the-fly use: as one must load pretrained SAE encoder + decoder for every steered layer, where each SAE block can be larger than the layer it modifies. Further, the patch is applied to every newly decoded token, thus risking divergence from the LRM’s original reasoning path — a concern we have discussed above under the L2S paragraph, given today’s limited reasoning benchmarks.

Last, we have \citet{mahaut2025repetitions} being a phenomenological/diagnostic study that analyzes how repetition arises via attention-head patterns but proposes no application-focused solutions. Its relationship to WSC is rather tangential, but we thought featuring here might interest the broader audiences.

\section{Details of Regeneration.}
\label{app:details}
We conduct all generation experiments on 4× NVIDIA A100 80G GPUs. During the rescue regeneration stage, we use \texttt{tensor\_parallel = 4} to fully leverage model parallelism across the available GPUs. 
\subsection{Initial Generation Settings}
We allow the model to generate up to 32k tokens during the initial decoding phase. This is consistent across all models and tasks. 
\subsection{Rescue Regeneration Settings}
we apply a fixed token budget during the rescue regeneration stage. Table \ref{tab:generation_hyper} summarizes the settings used in our experiments.
\begin{table}[ht]
\centering
\caption{Rescue regeneration budget (after chopping) for all experiments. (unit: \# of tokens)}
\resizebox{\linewidth}{!}{
\begin{tabular}{lc|c|c|c}

\toprule
 
Model   &    \textbf{GSM8K} &    \textbf{MATH-500} &    \textbf{AIME25} &    \textbf{GPQA-Diamond}  \\

\midrule

\multicolumn{5}{c}{\textbf{$\tau$ = 0.0}} \\

\midrule

Qwen-1.5B & 4k   &    4k     &    NA &    4k \\
Qwen-7B & 4k   &    4k     &    NA &    4k \\
Llama-8B  & 4k   &    4k     &    NA &    4k \\

\midrule

\multicolumn{5}{c}{\textbf{$\tau$ = 0.6}} \\

\midrule

Qwen-1.5B & 4k   &    4k     &    8k &    4k \\
Qwen-7B &  4k   &    4k     &    8k &    4k  \\
Llama-8B  & 4k   &    4k     &    4k &    4k \\

\bottomrule
\end{tabular}
}
\label{tab:generation_hyper}
\end{table}
\begin{tcolorbox}[colback=lightgray!10, colframe=black, title={Rescue Regeneration Prompt}]
I can find a clearer solution if I focus on the core problem.
\end{tcolorbox}
\section{Training Details of Linear Classifier}
\label{app:training details}
We train a single-layer logistic classifier with its default hyper-parameters: Adam optimizer (learning rate $1\times10^{-2}$, weight decay~$0$), \texttt{BCEWithLogitsLoss}, and a mini-batch size of $8\,192$.  
Training proceeds for $50$ epochs, with all random seeds fixed at~$41$ for reproducibility.  
To mitigate label imbalance, we first rebalance the training set to a 1:1 ratio of positive to negative chunks, and (where minor residual skew remains) set \texttt{pos\_weight} to the inverse class frequency.
\section{Details of Chopper}
\label{app:intervention_policy}
At each generation step we compute a repetition score $p_i$ and classify the current sentence as short or long based on its token count and the parameter \texttt{len\_threshold}. We maintain two counters: \texttt{long\_streak} for consecutive long sentences with $p_i>\texttt{thresh}$ and \texttt{short\_streak} for consecutive short sentences with $p_i>\texttt{thresh}$. Whenever $p_i\le\texttt{thresh}$ we reset the corresponding counter. We stop generation and trim all remaining sentences as soon as \texttt{long\_streak} reaches \texttt{streak\_len} or \texttt{short\_streak} reaches \texttt{short\_streak\_len}. In our experiments we set \texttt{thresh}=0.5, \texttt{streak\_len}=2, \texttt{len\_threshold}=10, and \texttt{short\_streak\_len}=5.



\section{Datasets Details}
\label{app:dataset_details}

\subsection{Linear Classifier Training Corpus}
\textbf{s1K}~\cite{muennighoff2025s1} contains 1 000 multi-domain, competition-style questions
(math, science, logic, general reasoning) with chain-of-thought solutions.
The dataset is released under the Apache~2.0 license.

\subsection{Evaluation Datasets.}
\begin{itemize}[leftmargin=*, noitemsep, topsep=0pt]
    \item \textbf{GSM8K}: 8792 grade-school word-problems  
          (7473 train / 1319 test) that each require 2–8 arithmetic steps. We use the 1,319-item test set.

    \item \textbf{MATH-500}: A 500-problem test subset drawn from the 12,500-item MATH dataset. We use this 500-item test set.
          \textsc{MATH} benchmark, covering algebra, number theory, geometry,
          combinatorics and precalculus with worked solutions.

    \item \textbf{GPQA-Diamond}: 198 multiple-choice
          graduate-level questions across physics, biology and chemistry designed
          to defeat information-retrieval baselines.

    \item \textbf{AIME25 (2025)}: 30 free-response problems (AIME I + II 2025)
          requiring creative high-school competition math; answers are
          three-digit integers.
\end{itemize}

\subsection{ Availability and Licensing of Artifacts}

\begin{itemize}[leftmargin=*, noitemsep, topsep=0pt]
    \item \textbf{Datasets}
    \begin{itemize}[leftmargin=*, noitemsep, topsep=0pt]
        \item \textbf{s1K}: Apache 2.0.  
        \item \textbf{GSM8K}: MIT.  
        \item \textbf{MATH-500}: MIT (inherits parent benchmark license).  
        \item \textbf{GPQA Diamond}: MIT.  
        \item \textbf{AIME25}: MIT license for the JSON wrapper; original problem
              statements © 2025 MAA, redistributed here under academic fair use.
    \end{itemize}

    \item \textbf{Models}
        \begin{itemize}[leftmargin=*, noitemsep, topsep=0pt]
        \item \textbf{DeepSeek-R1-Distill-Qwen-1.5B}, 
              \textbf{DeepSeek-R1-Distill-Qwen-7B}, and 
              \textbf{DeepSeek-R1-Distill-Llama-8B}:  
              All three checkpoints are released by DeepSeek under the \textbf{MIT License}, which permits commercial use, redistribution, and the creation of derivative works without additional approval.\footnote{See the model cards on HuggingFace for the exact license text.}
              Although each model is distilled from its respective parent (Qwen-2.5~\cite{Yang2024Qwen25TR} or Llama-3.1~\cite{grattafiori2024llama}), the redistributed weights themselves inherit the MIT terms.
    \end{itemize}
    \item \textbf{Code}  
          All custom scripts will be released under the MIT license.
\end{itemize}

All artifacts used in this work have been utilized in a manner consistent with their original intended use, as specified by their respective licenses. No proprietary or restricted data were included.
\section{WordSaladChopper Algorithm}
We present the pseudocode for WordSaladChopper in Algorithm~\ref{alg:wsc}.
\begin{algorithm}[t]
\caption{WordSaladChopper}
\label{alg:wsc}
\begin{small}
\begin{algorithmic}[1]
\State \textbf{Inputs:} $M$, $C$, $P$, $R$, $L$, $\texttt{params}$
\State $ids \gets \texttt{tokenize}(P)$
\State $long\_streak, short\_streak \gets 0, 0$
\State $last\_nl\_pos \gets |ids| - 1$
\While{$|ids| < L$}
    \State $logits, h \gets M.\texttt{forward}(ids)$
    \State $next\_id \gets \texttt{sample}(logits)$
    \State $ids.\texttt{append}(next\_id)$
    \If{$next\_id \in \texttt{NEWLINE\_TOKEN\_IDS}$}
        \State \textit{// repetition probability}
        \State $p \gets C(h)$
        \State $chunk\_len \gets |ids| - last\_nl\_pos - 1$
        \State $last\_nl\_pos \gets |ids| - 1$
        \State $is\_rep \gets (p > \texttt{thresh})$
        \If{$is\_rep$}
            \If{$chunk\_len \geq \texttt{len\_threshold}$}
                \State $long\_streak \gets long\_streak + 1$
                \State $short\_streak \gets 0$
            \Else
                \State $short\_streak \gets short\_streak + 1$
                \State $long\_streak \gets 0$
            \EndIf
        \Else
            \State $long\_streak, short\_streak \gets 0, 0$
        \EndIf
        \State $chop\_now \gets (long\_streak \geq \texttt{streak\_len})$
        \State \quad or $(short\_streak \geq \texttt{short\_streak\_len})$
        \If{$chop\_now$}
            \State \textit{// CHOP}
            \State $ids \gets ids[:-(chunk\_len + 1)]$
            \State \textit{// Append regeneration prompt}
            \State $ids.\texttt{extend}(\texttt{tokenize}(R))$
            \State \Return \texttt{continue\_until\_eos}$(M, ids, L)$
        \EndIf
    \EndIf
\EndWhile
\State \Return \texttt{detokenize}$(ids)$
\end{algorithmic}
\end{small}
\end{algorithm}

\section{Case Studies}
\label{sec:case-study}

We provide qualitative demonstrations of degeneration behaviors and our method's intervention strategy.

\paragraph{Case 1: Semantic Loop from Unresolved Ambiguity (MATH-500 \#462).}
The model begins with valid reasoning but then becomes trapped in a semantic loop --- repeating the same confusion without resolution:
\begin{quote}
\small
\texttt{``But when I added step-by-step, I got 9997.\textbackslash n\textbackslash n''} \\
\texttt{``But wait, 6270 + 3737 is 10,007, so why is the step-by-step adding 3000, 700, 30, and 7 giving me 9997?\textbackslash n\textbackslash n''} \\
\texttt{``But why does the step-by-step addition give me 9997?\textbackslash n\textbackslash n''} (chopped here) \\
\texttt{``Wait, so 6270 + 3737 is 10,007...\textbackslash n\textbackslash n''}
\end{quote}

WSC detects early signs of degeneration and chops at the third chunk, followed by a regeneration prompt. The regenerated continuation quickly resolves the problem with correct reasoning within a 4k budget.

\paragraph{Case 2: Endless Enumeration without Convergence (MATH-500 \#110).}
The model attempts a brute-force enumeration without reaching a conclusion:
\begin{quote}
\small
\texttt{``For k=1: ...''} \\
\texttt{``k=12: ...''} \\
\texttt{``k=14: ...''} (chopped here) \\
\texttt{``k=27: ...''}
\end{quote}

Here, WSC intervenes at chunk 318 to prevent further unbounded enumeration, ensuring the continuation remains within budget. This illustrates WSC’s ability to detect degeneration early and prevent catastrophic repetition.

\section{Discussion on Choice of Delimiter}\label{app:delimiter}

A natural question concerns our use of ``\texttt{\textbackslash n\textbackslash n}'' as the segmentation point for reasoning traces. We provide both intuition and empirical evidence for this choice.

\paragraph{Rationale} 
We opt for ``\texttt{\textbackslash n\textbackslash n}'' because it is (i) \emph{prevalent} in the reasoning traces of Large Reasoning Models (LRMs), and (ii) carries \emph{minimal semantic meaning}. In contrast, tokens such as ``Wait'' or ``Alternatively'' embed semantic cues that may bias downstream classifiers. While there is no universally agreed delimiter for LRMs due to their recency, choosing minimal or non-semantic trailing tokens as chunk representatives has long been a practice in NLP. For example, dense retrievers often use the \texttt{<eos>} token at the end of a passage as the vector representation of the whole passage~\cite{wang-etal-2024-improving-text}, and efficiency works register special \texttt{<beacon>} tokens at chunk boundaries to encode chunk-level information~\cite{zhang2025long}. The ``\texttt{\textbackslash n\textbackslash n}'' token naturally fulfills both criteria (minimal / non-semantic + trailing), making it a strong candidate for our purposes.

\paragraph{Empirical Evidence} 
We further observe that sentences with similar semantic content yield different classifier scores at their trailing ``\texttt{\textbackslash n\textbackslash n}''. As repetitions accumulate, later chunks become progressively easier for the classifier to identify as degenerate. Table~\ref{tab:delimiter_case} illustrates this progression: classifier scores (0 -- 1, with higher scores indicating stronger repetition) sharply increase with more repetitions, making ``\texttt{\textbackslash n\textbackslash n}'' an effective marker for repetition detection.

\begin{table}[h]
\centering
\caption{Classifier scores at the trailing ``\texttt{\textbackslash n\textbackslash n}'' across repetitions 
(MATH-500 \#462, DeepSeek-R1-Distill-Qwen-7B, Temp=0.6).}
\label{tab:delimiter_case}
\resizebox{\linewidth}{!}{
\begin{tabular}{clc}
\toprule
Chunk idx & Sentence & Score \\
\midrule
209 & ``But when I added step-by-step, I got 9997.\textbackslash n\textbackslash n'' & 1.19e-10 \\
255 & ``But when I did the step-by-step addition, I got 9997.\textbackslash n\textbackslash n'' & 3.69e-5 \\
$\cdots$ & $\cdots$ & $\cdots$ \\
430 & ``Wait, so that must mean that 6270 + 3737 is 9997.\textbackslash n\textbackslash n'' & 1.000 \\
\bottomrule
\end{tabular}
}
\end{table}
\paragraph{Takeaway} 
These results demonstrate that ``\texttt{\textbackslash n\textbackslash n}'' provides both a theoretically sound and empirically effective delimiter for identifying the onset of repetitive behavior in LRMs. It strikes a balance between being common in generation, semantically neutral, and progressively sensitive to degenerative repetition patterns.

\section{Latency of On-the-fly Detector and its Integration Strategies}\label{app:latency}
\paragraph{Takeaway}
Our linear classifier for word-salad detection can be integrated into LRM decoding with \emph{negligible to near-zero} latency overhead.
When implemented asynchronously (in parallel with the LLM forward pass), it introduces effectively no extra wall-clock latency.
When implemented sequentially (LLM waits for the classifier at each \texttt{\textbackslash n\textbackslash n}), the overhead is bounded to roughly \textbf{0 -- 0.4\%} under our settings.

\subsection{Integration Strategies}
\paragraph{Asynchronous (parallel) integration}
Once an \texttt{\textbackslash n\textbackslash n} token is generated, we extract its hidden state and run the linear classifier \emph{in parallel} with the next LLM forward.
Because a single LLM forward step is consistently slower than a single classifier forward, the classifier latency is fully hidden.
This mode adds \emph{practically no} additional latency.  

\paragraph{Sequential (wait-on-classifier) integration}
Alternatively, the LLM may \emph{wait} for the classifier decision at each \texttt{\textbackslash n\textbackslash n} before proceeding.
In that case, the overhead equals one classifier forward per reasoning chunk.
Based on the runtimes in Table~\ref{tab:latency_runtime} and an average chunk length of $\sim$32 tokens on MATH-500, this corresponds to an estimated overhead of about \textbf{0.4\%} per chunk for a 7B model.

\subsection{Empirical Runtime}
We benchmark the latency of a one-token LLM forward pass versus a single classifier prediction using the hidden state of the trailing \texttt{\textbackslash n\textbackslash n}. 
The classifier inference is consistently $\sim$5\,ms, significantly faster than an LLM forward step.

\begin{table}[h]
\centering
\caption{Average runtime over 5 runs. ``LLM Fwd'' = one-token forward; ``Clf Fwd'' = one classifier prediction from the trailing hidden state.}
\label{tab:latency_runtime}
\resizebox{\linewidth}{!}{
\begin{tabular}{lccc}
\toprule
Model & LLM Fwd (1 tok) & Clf Fwd (1 pred.) & Hidden Dim \\
\midrule
DeepSeek-R1-Distill-Qwen-1.5B & 31.52 ms & 4.96 ms & 1536 \\
DeepSeek-R1-Distill-Qwen-7B   & 39.16 ms & 4.95 ms & 3584 \\
DeepSeek-R1-Distill-Llama-8B  & 41.12 ms & 4.95 ms & 4096 \\
\bottomrule
\end{tabular}}
\end{table}


\subsection{Overhead Analysis}
Let $T_{\text{LLM}}$ and $T_{\text{clf}}$ denote the per-step runtime of the LLM and the classifier, respectively, and let $\bar{L}$ be the average chunk length (in tokens).
Under the sequential mode, the per-chunk overhead ratio is
\[
\textstyle \frac{T_{\text{clf}}}{\bar{L}\cdot T_{\text{LLM}}}\, .
\]
With $T_{\text{LLM}}\approx39.16\,\text{ms}$, $T_{\text{clf}}\approx4.95\,\text{ms}$, and $\bar{L}\approx32$, the estimated overhead is
\[
\frac{4.95}{32 \times 39.16} \approx 0.004 \;=\; 0.4\% \, .
\]
This is a theoretical estimate rather than an end-to-end measurement.
\section{Additional Results on Qwen3}
\label{app:qwen3}

\paragraph{Setup}
To assess generalization beyond DeepSeek-R1 models, we evaluate the WordSaladChopper (WSC) classifier on \textbf{Qwen3-8B}~\cite{yang2025qwen3} in the \emph{thinking} mode across three benchmarks (GSM8K, MATH-500, AIME25) and two decoding temperatures (0.0, 0.6). The classifier operates on the hidden state of the trailing ``\texttt{\textbackslash n\textbackslash n}'' token to detect repetitive (``word salad'') chunks on-the-fly.%

\begin{table}[h!]
  \centering
  \small
  \setlength{\tabcolsep}{6pt}
  \caption{Classifier accuracy (\%) for word-salad chunk detection on \textbf{Qwen3-8B}. Higher is better.}
  \label{tab:qwen3-acc}
  \begin{tabular}{lccc}
    \toprule
    \textbf{Temp} & \textbf{GSM8K} & \textbf{MATH-500} & \textbf{AIME'25} \\
    \midrule
    0.0 & 78.0 & 88.1 & 81.4 \\
    0.6 & 78.9 & 87.0 & 84.3 \\
    \bottomrule
  \end{tabular}
\end{table}

\paragraph{Findings}
As shown in Table~\ref{tab:qwen3-acc}, the classifier achieves robust accuracy on Qwen3-8B, averaging around \(\sim\)83\% across datasets/temperatures. This is lower than on DeepSeek-R1-Distill-Qwen-7B (e.g., 92.72/92.31/89.77 at $\tau=0.0$), but remains usable in practice since WSC triggers a chop only after multiple consecutive detections, and simple gating rules can further reduce unnecessary interventions in hybrid reasoning pipelines.  

\end{document}